\documentclass[letterpaper]{article} 
\usepackage{aaai23}  
\usepackage{times}  
\usepackage{helvet}  
\usepackage{courier}  
\usepackage[hyphens]{url}  
\usepackage{graphicx} 
\usepackage{natbib}  
\usepackage{caption} 
\usepackage{xspace}												
\usepackage[usenames,dvipsnames,svgnames,table]{xcolor}			
\usepackage{mathtools}                                          
\usepackage{amssymb}											
\usepackage{trimclip}                                           
\usepackage{amsmath}
\usepackage{multirow}
\usepackage{booktabs}
\usepackage[inline,shortlabels]{enumitem}
\usepackage{microtype}
\usepackage[none]{hyphenat}

\usepackage{algorithm}
\usepackage[noend]{algorithmic}

\renewcommand{\k}[1]{\textit{#1}}

\newcommand{\mk}[1]{\mathit{#1}}
\renewcommand{\b}[1]{\textbf{#1}}
\newcommand{\Q}[1]{\textbf{(Q#1)}}
\newcommand{\Step}[1]{\textbf{(S#1)}}

\newcommand{\A}{\mathcal{A}}

\newcommand{\M}{\mathcal{M}}

\newcommand{\F}{\mathcal{F}}
\renewcommand{\L}{\mathcal{L}}

\renewcommand{\S}{\mathcal{S}}

\newcommand{\<}{\langle}
\renewcommand{\>}{\rangle}
\newcommand{\into}{\mapsto}
\newcommand{\dotrightarrow}{\clipbox{0pt 0 {.675\width} 
0}{\ensuremath{\rightarrow}} \makebox[1.1pt]{} {\cdot} {\cdot} {\cdot} 
\makebox[1.1pt]{} \clipbox{{.55\width} 0 0pt -1pt}{\ensuremath{\rightarrow}}}
\newcommand{\x}{\times}

\newcommand{\DCS}{{\scshape otf-dcs}\xspace}
\newcommand{\RL}{{\scshape rl}\xspace}
\newcommand{\RLNS}{{\scshape rlns}\xspace}
\newcommand{\RA}{{\scshape ra}\xspace}
\newcommand{\Random}{{\scshape random}\xspace}

\renewcommand{\l}{\ell}
\newcommand{\init}[1]{\bar{#1}} 

\newcommand{\D}{\rightarrow}
\newcommand{\step}[2]{\overset{#1}{\D}_{#2}}
\newcommand{\runw}[2]{\,\overset{#1}{\dotrightarrow}_{#2}\,}
\newcommand{\structure}{\mk{A}}
\newcommand{\winningstates}{\mk{WS}}
\newcommand{\losingstates}{\mk{LS}}

\title{Exploration Policies for On-the-Fly Controller Synthesis: \\ A Reinforcement Learning Approach}
\author{
    Tomás Delgado\textsuperscript{\rm 1}, 
    Marco Sánchez Sorondo\textsuperscript{\rm 1}, 
    Víctor Braberman\textsuperscript{\rm 1}, 
    Sebastián Uchitel\textsuperscript{\rm 1}\textsuperscript{\rm 2}
}
\affiliations{
    \textsuperscript{\rm 1}Universidad de Buenos Aires\\
    \textsuperscript{\rm 2}Imperial College\\
    tdelgado@dc.uba.ar, msorondo@dc.uba.ar, vbraber@dc.uba.ar, suchitel@dc.uba.ar
}

\begin{document}

\maketitle

\begin{abstract}

Controller synthesis is in essence a case of model-based planning for non-deterministic environments in which plans (actually “strategies”) are meant to preserve system goals indefinitely. In the case of supervisory control environments are specified as the parallel composition of state machines and valid strategies are required to be “non-blocking” (i.e., always enabling the environment to reach certain marked states) in addition to safe (i.e., keep the system within a safe zone). Recently, On-the-fly Directed Controller Synthesis techniques were proposed to avoid the exploration of the entire -and exponentially large- environment space, at the cost of non-maximal permissiveness, to either find a strategy or conclude that there is none. The incremental exploration of the plant is currently guided by a domain-independent human-designed heuristic.

In this work, we propose a new method for obtaining heuristics based on Reinforcement Learning (RL). The synthesis algorithm is thus framed as an RL task with an unbounded action space and a modified version of DQN is used. With a simple and general set of features that abstracts both states and actions, we show that it is possible to learn heuristics on small versions of a problem that generalize to the larger instances, effectively doing zero-shot policy transfer. Our agents learn from scratch in a highly partially observable RL task and outperform the existing heuristic overall, in instances unseen during training.

\end{abstract}

\section{Introduction}

Reactive systems in domains such as communication networks, automated manufacturing, air traffic control, and robotics, can benefit from the automated construction of correct control strategies. Discrete Event Control~\citep{Wonham:1987:SCS}, Automated Planning \citep{Nau:2004:AP} and Reactive Synthesis \citep{Pnueli:1989:RS} are fields that address this automated construction problem. 
Although they have representational, expressiveness and algorithmic differences, the three 
must deal with the state explosion that results from analysing a compact input description with exponentially large underlying semantics. Recently, an increasing number of studies aimed at relating the fields \citep[e.g.][]{Ehlers16,DBLP:conf/ijcai/SardinaD15, Camacho_Bienvenu_McIlraith_2021, DBLP:conf/aaai/HoffmannHKSKM20}.

In this paper we study an approach to discrete event systems (DES) control in which a \k{plant} to be controlled is specified modularly as the parallel composition of communicating finite state automata \citep{Wonham:1988:MSC}. The aim is to build a \k{director} that is \k{safe} and \k{nonblocking}. That is, it should guarantee that a marked state of the plant can always be reached while ensuring that unsafe plant states are never reached. Directors~\citep{Huang:2008:DCD} are controllers that enable at most one controllable event at any time, in contrast to \k{supervisors}, which are maximally permissive.

Composing the automata of the plant can result in an exponential state explosion. Approaches that first build the full plant and compute a director can fail within a time and memory budget even when there is a director that keeps the system in a very small proportion of the full plant state space.  
On-the-fly Directed Controller Synthesis (\DCS) \citep{CiolekDCS} attempts to avoid state explosion by exploring the composed plant incrementally, checking for the existence of directors after each new transition is added. 
If guided by good heuristics, this process allows finding controllers by building only the parts of the plant that the controllers themselves enable reaching. In the same work, several manually designed heuristics were proposed and it was shown that for certain domains it is possible to solve instances that cannot be solved by a fully-compose and synthesise approach.

\k{In this work we propose replacing manually designed heuristics for} \DCS \k{with an exploration policy learned via Reinforcement Learning (RL) that is efficient for large instances of a parameterized control problem having been trained only on small instances of the problem}.
Note that we cannot use RL to learn the control strategy itself because RL cannot provide full guarantees that the controller will satisfy the safety and non-blocking properties. We use RL to learn an exploration strategy that minimizes the number of transitions to be added to the plant by the \DCS algorithm. Additionally, note that RL has traditionally focused on tasks where agents are evaluated in the same environment as they are trained. This is not useful in our application because we are ultimately interested in solving large instances that cannot currently be solved in reasonable amounts of time.
We use RL to find policies that \textit{generalize} to unseen instances.

For this purpose, the \DCS algorithm (and not the control problem being solved) is framed as a Markov Decision Process in which reward is obtained by minimizing the number of explored transitions of the plant. This task can be challenging because
(1) it has a sparse reward (``there is no information about whether one action sequence is better than another until it terminates'', \citet{planningRL}), 
(2) the action set is large (one per plant transition) and actions are used only once per episode, and (3) the state has a complex graph structure. 

We use a modified version of DQN~\citep{mnih2013playing} and we \k{abstract} both actions and states to a feature space that is unique for all instances of a parameterized control problem. This addresses challenges (2) and (3), and allows for generalization, but causes the RL task to be highly partially observable due to the information loss of the set of features. We propose training in small instances of a problem (partially addressing challenge (1)) for a relatively short time with a small neural network. Then, a uniform sample of the policies obtained during training is tested on slightly larger instances. The one that shows best generalization capabilities is selected and then fully evaluated in a large time-out setting.

Our results show first that with this technique it is possible to learn competitive heuristics on the training instances; and second, that these policies are effective when used in larger instances. Our agents are evaluated both in terms of expanded transitions and in terms of solved instances within a time budget, and overall they outperform the best heuristic from \citet{CiolekDCS}, pushing the boundaries of instances solved in various of the benchmark problems.

\section{Background}

\subsection{Modular Directed Control}

The discrete event system (DES) or plant to be controlled is defined as a tuple $E = (S_E, A_E, \D_E, \init{s}, M_E)$, where $S_E$ is a finite set of states; $A_E$ is a finite set of event labels, partitioned as $A_E^C \cup A_E^U$, the controllable and uncontrollable events; $\D_E : S_E \x A_E  \into S_E$ is a partial function that encodes the transitions; $\init{s} \in S_E$ is the initial state; and $M_E \subseteq S_E$ is a set of marked states.

This automaton defines a language $\L(E) \subseteq A_E^*$, where $^*$ denotes the Kleene closure in the usual manner. A word $w \in A_E^*$ belongs to the language if it follows $\D_E$ with a sequence of states $\init{s}=s_0 \ldots s_t$. In this case we note $\init{s} \runw{w}{E} s_t$.

A function that based on the observed behaviour of a plant decides which controllable events are allowed is referred to as a control function. Given a DES $E$, a controller is a function $\sigma : A_E^* \into \mathcal{P}(A_E^C)$. A word $w \in \L(E)$ belongs to $\L^\sigma(E)$, the language generated by $\sigma$, if each event $l_i$ is either uncontrollable or enabled by $\sigma(l_0 \ldots l_{i-1})$.

Given a plant, we wish to find a controller that ensures that a marked state can be reached from any reachable state (even from marked states). The \k{non-blocking} property captures this idea. Formally, a controller $\sigma$ for a given DES is non-blocking if for any trace $w \in \L^\sigma(E)$, there is a non-empty word $w' \in A_E^*$ such that the concatenation $ww' \in L^\sigma(E)$ and $\init{s} \runw{\;ww'}{E} s_m$ for some $s_m \in M_E$. Additionally, a controller is a \k{director} if $|\sigma(w)| \leq 1$ for all $w \in A_E^*$.

Note that with this definition a non-blocking controller must also be \textit{safe} in the sense that it cannot allow a deadlock state to be reachable (i.e. a state with no outgoing transitions). Unsafe states can be modelled as deadlock states.

Modular modelling of DES control problems~\citep{Ramadge:1989:SC} supports 
describing the plant by means of multiple deterministic automata and their synchronous 
product or \k{parallel composition}.

The parallel composition $(\|)$ of two DES $T$ and $Q$ yields a DES $T \| Q = (S_T 
{\times} S_Q, A_T {\cup} 
A_Q, \D_{T\|Q}, \<\init{t},\init{q}\>, M_T {\times} M_Q)$, where $A_{T \| Q}^C$ = $A_T^C \cup A_Q^C$ and $\D_{T\|Q}$ is the smallest relation that satisfies the following rules:
\begin{enumerate}[(i)]
    \item if $t \step{\l}{T} t'$ and  $\l \in A_{T} {\setminus} A_{Q}$ then $\<t,q\> {\step{\l}{T\|Q}} \<t'\!,\!q\>$,
    \item if $q\step{\l}{Q} q'$ and  $\l \in A_{Q} {\setminus} A_{T}$ then $\<t,q\> {\step{\l}{T\|Q}} \<t\!,\!q'\>$,
    \item if $t \step{\l}{T} t'$, $q \step{\l}{Q} q'$, and  $\l \in A_{T} {\cap} A_{Q}$ then $\<t,q\> \step{\l}{T\|Q} \<t',q'\>$.
\end{enumerate}

A Modular Directed Control Problem, or simply \textit{control problem} in this paper, is given by a set of deterministic automata $E = (E^1, \ldots, E^n)$. A solution to this problem is a non-blocking director for $E^1\|\ldots\|E^n$. Additionally, the control problems that we aim to solve in this work are parametric. A \k{control problem domain} is a set of instances $\Pi = \{E_p : p \in \mathcal{C}\}$, where each $E_p$ is a (modular) control problem and $\mathcal{C}$ is a set of possible parameters. In our case, control problems in each domain are generated by the same specification, which takes parameters $p$ as input and is written in the FSP language \citep{magee2014concurrency}.

\subsection{On-the-Fly Modular Directed Control}

The Modular Directed Control Problem can be solved by fully building the composed plant and running a monolithic algorithm such as the presented by \citet{Huang:2008:DCD}. While this quickly becomes intractable, there are problems for which the state explosion can be delayed significantly by exploring a small subset of the plant that is enough to determine a control strategy (or to conclude that there is none). The \DCS algorithm \citep{CiolekDCS} is briefly summarized in Algorithm~\ref{alg:dcs}. It performs a best-first search of the composed plant, adding one transition at a time from the exploration frontier to a partial exploration structure ($\structure$). 

Formally, given $E = (S_E, A_E, \D_E, \init{s}, M_E)$, a control problem, and $h = \{a_0, \ldots, a_t\} \subseteq \D_E$, a sequence of transitions, the \k{exploration frontier} of $E$ after expanding sequence $h$ is $\F(E, h)$, the set of transitions $(s, \l, s') \in (\D_E \setminus \hspace{0.1cm} h)$ such that $s = \init{s}$ or $(s'', \l, s) \in h$ for some $s''$. An \k{exploration sequence} for $E$ is $\{a_0, \ldots, a_t\} \subseteq \D_E$ such that $a_i \in \F(E, \{a_0, \ldots, a_{i-1}\})$ for $0 \leq i \leq t$.

\begin{algorithm}[t!]
\caption{On-the-fly exploration procedure.}\label{alg:dcs}
Input: $E^i$ $=$ $(S_{E^i}, A_{E^i}, \D_{E^i}, \init{s}^i, M_{E^i})$, components of $E$, and a heuristic $H$ for $E$.
\begin{algorithmic}
\STATE $\init{s} \gets (\init{s}^1,\ldots,\init{s}^n)$
\STATE $h \gets$ Empty list.
\STATE $\structure \gets (\{\init{s}\}, A_E, \emptyset, \init{s}, M_E \cap \{\init{s}\})$
\STATE $\winningstates \gets \emptyset$
\STATE $\losingstates \gets \emptyset$
\WHILE{$\init{s} \not\in \winningstates \cup \losingstates$}
  \STATE $a \gets $ action selected from $\F(E, h)$ using $H$.
  \STATE $expandAndPropagate(a, \structure, \winningstates, \losingstates)$
  \STATE Append $a$ to $h$
\ENDWHILE
\IF{$\init{s} \in \winningstates$}
  \STATE return $buildController(h, \winningstates)$
\ELSE
  \STATE return $unrealizable$
\ENDIF
\end{algorithmic}
\end{algorithm}

With each added transition, $expandAndPropagate$ updates the classification of states in sets of losing ($\losingstates$), winning ($\winningstates$), or neither. We say that a state $s \in E$ is \k{winning} (resp. \k{losing}) in a plant $E$ if there is a (resp. there is no) solution for $E_s$, where $E_s$ is the result of changing the initial state of $E$ to $s$. Essentially, a state will be winning if it is part of a loop that has a marked state and that has no uncontrollable events that go to states outside the loop, or if it can controllably reach a winning state. A state will be losing if it has no path to a marked state, if it can be forced by uncontrollable events towards a losing state, or if all its events are controllable but they lead to losing states. For a given (uncompleted) exploration sequence a state is defined to be winning (resp. losing) if it is winning (resp. losing) when assuming that every transition in the exploration frontier goes to a losing (resp. winning) state. Note that it is possible for a state to be neither winning nor losing when the plant is not completely explored.

A key remark is that the classification performed by the algorithm is correct and complete (no false positives or false negatives). Hence, a verdict for the initial state will be found in worst case after the last transition is expanded, and that verdict is guaranteed to be correct. A heuristic, then, will try to minimize the number of transitions explored, but even with very poor decisions, the completeness of the synthesis algorithm is not threatened. In this work, heuristics are replaced by learned exploration policies that evaluate transitions observing a set of features that are computed throughout the algorithm.


\subsection{Q-Learning with Function Approximation}

Reinforcement Learning considers an agent that interacts iteratively with its environment, learning to maximize a reward function. An episodic RL task can be formalized as a Markov Decision Process (MDP) $\M = (\S, \A, P, r, S_0)$ where $\S$ is a set of states; $\A$ is a set of actions; $P : \S \x \A \x \S \into [0, 1]$ encodes the probability $Pr\{s'|a, s\}$ of observing state $s'$ after selecting action $a$ in state $s$; $r : \S \x \A \x \S \into \mathbb{R}$ is a reward function; and $S_0$ is an initial state. The set of available actions in state $s$ is denoted by $\A(s)$. Then, the goal is to find a policy $\pi : \S \x \A \into [0, 1]$ that maximises the expected accumulated reward $\mathbb{E}_\pi[\sum_{t=0}^T R_t]$, where $T$ and $R_t$ are random variables describing the number of steps and the reward at step $t$ for an episode.

Q-Learning \citep{Watkins:1992} approximates an optimal action-value function $Q^*: \S \x \A \into \mathbb{R}$, which can be defined as $Q^*(s, a) = \max\limits_\pi \mathbb{E}_\pi[\sum_{t'=t}^T R_{t'} | S_t=s, A_t = a]$.
That is, the expected accumulated reward that is obtained after taking action $a$ in state $s$ and then following an optimal policy. Any action-value function directly induces a greedy policy that always chooses the action that maximizes $Q$ for the given state, and the policy induced by $Q^*$ is an optimal policy. When the state-action space is intractable, the tabular form for $Q$ is commonly replaced by a function approximator $\hat{Q}_w(s, a)$. Then, stochastic gradient descent updates on $w$ can be used to minimize the one-step error $(R_{t+1} + \max_{a \in A_{S_{t+1}}} \hat{Q}_w(S_{t+1}, a) - \hat{Q}_w(S_t, A_t))^2$.

In the tabular case convergence is guaranteed under sufficient exploration of the state-action space and the step-size parameter being reduced appropriately. However, it is known that using function approximation together with this off-policy one-step error (sometimes called a \k{deadly triad}) can cause divergence (see Chapter 11 of \citet{suttonbarto} for instructive examples). Two techniques that have been proposed to restore stability are Experience Replay and Fixed Q-Targets \citep{mnih2013playing,lin1992reinforcement}. Instead of updating directly from each observation with the current value function, a minibatch update is used with a uniform sample of the last $B$ experiences, towards a fixed $Q$ function that is updated periodically.

\section{\DCS Using Reinforcement Learning}

\subsection{Exploration Optimization as an RL Task}

In this section we define an MDP that, although not immediately practical, exactly represents the problem of minimizing exploration. Given the similarity between MDPs and DES, it is tempting to define an MDP in which states and actions in the DES correspond to states and actions in the MDP, and a reward in the MDP is given when a marked state is visited in the DES. However, with this MDP the resulting policy would be able to select an event in a plant state, and not a transition in the exploration frontier. Also, it is not obvious how to encode uncontrollable transitions, since it can be crucial to select them in the exploration problem, but in the control problem they are out of the control of the agent.

In our MDP, a state is defined as the state of the exploration process (the sequence of expanded transitions), and an action represents the expansion of a transition (a DES state-event pair). 
The dynamics ($P$) are simply given by adding transitions to the sequence, and the rewards are always $-1$. A terminal state is an exploration state in which the initial DES state is marked as winning or losing by \DCS.

More formally, given a control problem $E = E^1 \| \ldots \| E^n$, we define the associated MDP as $(\S, \A, P, r, S_0)$, where
\begin{itemize}
    \item $\S = \{h : h $ \ is an exploration sequence for $E \}$;
    \item $\A = S_E \x A_E$, $\A(s) = \{(s, \l) : \exists (s, \l, s') \in F(E, S_E) \}$;
    \item $P(s' | s, a) = 1$ if $a \in \F(E, s)$ and $s' = s a$, and 0 otherwise;
    \item $r(s, a, s') = -1 \ \forall s, a, s'$;
    \item $s \in \S$ is a terminal state if the initial state is winning or losing after expanding sequence $s$ in $E$;
    \item $S_0 = \emptyset$ (the empty sequence).
\end{itemize}

A positive property of this MDP is that it is an exact representation of our problem: a policy with reward $-R$ maps directly to a heuristic that expands $R$ transitions in the \DCS algorithm. The problem is that the state and action signals are completely impractical. First, the state is a sequence of explored transitions conforming a graph, which cannot be processed by traditional neural networks with a fixed input size. Second, the action space is large and only a variable-size subset of the actions is available at every step (the frontier). Note that in this MDP actions are taken at most once in a given episode.

Even more problems arise since, as will be further discussed in the next subsections, we want a learned exploration policy to be well-defined in larger instances of a domain. Plant states in different instances are different (because the corresponding plants are the composition of different sets of automata) and labels of events usually also change, because they can reference individual automata. Furthermore, the number of actions grows unbounded as the problems in a domain grow in size.

\subsection{Abstracting the Exploration State}

To solve the problems above, we propose abstracting the explored subgraph of the plant and the transitions available in the frontier, describing them with a general set of features $\phi(s, a) = (\phi_1(s, a), \ldots, \phi_{d_E}(s, a)) \in \mathbb{R}^{d_E}$. Then, agents observe in each state a list of feature vectors, one for each transition available in the frontier (the features that describe the state of the exploration are replicated).

This featurization has several advantages. First, the feature space has a fixed size ($d_E$), allowing the use of a traditional neural network. Second, it can make learning easier by simplifying the state signal and enriching the action signal; in particular, it allows generalizing across states and actions with similar features, which would not be possible with a granular unstructured identification. Third, and maybe most importantly, if agents learn in a feature space that is unique for a set of instances, the policy learned in one instance induces policies in all the others.

Nevertheless, the featurization may (and in our case will) introduce partial observability in both states and actions. Having partially observable states makes the task non-Markovian and can be modelled as a POMDP \citep{QLPOMDP}. However, our actions are also partially observable: the featurization does not need to fully characterize a transition in the plant. In the case of having two transitions in the frontier with the same feature vectors, the agent cannot distinguish them and one must be chosen arbitrarily (in this paper we choose the transition that entered the frontier first). If learning in this context is possible, the quality of the learned exploration policies will depend on the ability of the features to separate good state-action pairs from bad state-action pairs. Furthermore, the quality of the induced policies will only be preserved across instances if state-action pairs of different instances with similar features are similarly good (in terms of the number of transitions that can be expanded through them).

\subsection{Learning Algorithm}
\label{sec:algorithm}

In this section we describe how neural network-based Q-Learning can be used to solve our RL task. The algorithm used is essentially DQN \citep{mnih2013playing}. However, DQN relies on a fixed (relatively small) action set which, as it was discussed in the previous subsections, is not the case for our task. Thus, instead of using an architecture with one output for each action, we evaluate each action separately using a neural network with a single output. The input of the network is the feature vector $\phi(s, a) \in \mathbb{R}^{d_E}$ for each state-action pair $(s, a) \in \S \times \A$. The network estimates the optimal value function $Q^*(s, a)$ via $Q_w(\phi(s, a))$, where $w$ is a set of weights, and the Q-Learning update rule is used. Formally, this can be viewed as doing Q-Learning with function approximation, with the composition $Q \circ \phi$ as function approximator, and thus we have no theoretical guarantees of convergence.

Note that since each transition expanded has a reward of $-1$ and $Q$ is the expected sum of the rewards, the $Q$-values will be estimates of the (negative) expected number of transitions that will need to be expanded to finish the task after expanding transition $a$ in state $s$.

\begin{algorithm}[tb]
\caption{Q-Learning with function approximation for the Modular Directed Control RL task.}
\label{alg:learningalgorithm}
\textbf{Input}: A control problem $E$.
\begin{algorithmic}
\STATE $Env \gets$ \DCS solver environment for $E$.
\STATE Initialize $Q$ with random weights and input dimension $d_E$.
\STATE Initialize $Q'$ as a copy of $Q$.
\STATE Initialize buffer $B$ with observations from a random policy.
\STATE $S_0 \gets$ $reset(Env)$.
\FOR{$t=0$ \textbf{to} $T$}
\STATE $a_t \gets
  \begin{cases}
        \text{a random action}     & \text{with probability $\varepsilon$} \\
        arg\max\limits_{a \in \A(S_t)} Q(\phi(S_t, a)) & \text{otherwise}
  \end{cases}$
\STATE $S_{t+1} \gets$ Expand and propagate $a_t$ at $Env$.
\STATE Add $(\phi(S_t, a_t), S_{t+1})$ to $B$.
\STATE Sample transitions $(\phi(S_j, a_j), S_{j+1})$  randomly from $B$
\STATE $\delta_j \gets -1 + 
  \begin{cases}
        0 \hskip2.6cm \text{if $S_{j+1}$ is terminal} \\
        \max\limits_{a \in \A(S_{j+1})} Q'(\phi(S_{j+1}, a)) \hskip0.3cm \text{otherwise} \\
  \end{cases}$
\STATE Gradient descent on $Q$ with minibatch $(\phi(S_j, a_j), \delta_j)$
\STATE $Q' \gets Q$ if a fixed number of steps has passed.
\STATE $S_{t+1} \gets reset(Env)$ if $S_{t+1}$ is terminal.
\ENDFOR
\end{algorithmic}
\end{algorithm}

A pseudocode for the learning algorithm is shown in Algorithm~\ref{alg:learningalgorithm}. 
The agent synthesizes the same problem repeatedly until the time steps run out. At each step $t$, the feature vector $\phi(S_t, a)$ of each transition $a$ in the exploration frontier is evaluated using $Q$, and an $\varepsilon$-greedy action is selected. The environment propagates the verdicts of winning and losing states in the explored plant and the new experience is added to $B$, removing the oldest experience if necessary. At an implementation level, a vector of feature vectors, one for each transition $a \in \A(S_{t+1})$, is saved instead of $S_{t+1}$. After every step, a minibatch update is performed on $Q$ from a random sample of $B$. The target value $\delta_j$ for experience $(\phi(S_j, a_j), S_{j+1})$ is, if $S_{j+1}$ is not terminal, the value of the best feature vector in $S_{j+1}$, according to $Q'$, minus one (the reward). The target function is updated with a fixed frequency as a new copy of $Q$. Finally, if the new state is terminal the synthesis process is restarted.

Asymptotically, the evaluation of the neural network does not induce an overhead since the complexity of each iteration of \DCS (expanding a transition and propagating the verdicts) is bounded by $O(|S_{ES}|^2 \times |A_E|)$ and the number of transitions in the frontier is bounded by $O(|S_{ES}|^2)$. In practice, the worst-case bound for the propagation procedure could be reached rarely, and the evaluation of a large neural network could add a significant overhead. 

\subsection{Generalizing to Larger Instances}
\label{sec:endtoend}

Our work in this paper is concerned with scaling the synthesis procedure to large environments. Specifically, given a control problem domain $\Pi$, we want to find exploration policies that allow solving instances that cannot currently be solved using a reasonable amount of resources (time or memory). Since training in RL involves playing episodes repeatedly, our algorithmic design has an important constraint: \k{training cannot be performed in the instances that we want to solve}. Thus, one way forward is to find a methodology that leverages what can be learned in relatively small instances of a domain and attempts to use similar exploration strategies in the larger versions.

Clearly, for this to be possible, the larger versions should be related in some way to the training instances. This \k{homogeneity hypothesis} in our case is based on the fact that all instances $E_p \in \Pi$ are defined using the same parametric specification in the FSP language.

Since our learning algorithm presumably produces good exploration policies for a given instance, and the $Q$ functions that it generates induce policies in all instances of the same domain, our approach for solving the largest possible instances of a given problem $\Pi$ consists of three steps:
\begin{enumerate}
  \item[\Step{1}] Training in an instance $E_{p_0}$ (as described in section \ref{sec:algorithm}), saving $N$ agents sampled uniformly from the training process.
  \item[\Step{2}] Testing the policies obtained during \Step{1} on each instance $E_p \in \Pi$ with a small budget of transitions. The policy that generalized best (i.e. solved the most instances, breaking ties with total expanded transitions) is selected.
  \item[\Step{3}] The policy selected in \Step{2} is used with a full budget to solve as many instances as possible from $\Pi$.
\end{enumerate}

Although our hypothesis of homogeneity suggests that good performances in the training instance correlate in some hard-to-specify way to good performances in larger instances, it is clearly possible for a given set of weights to be an exception to this idea. A policy could be overfitted to the training instance in two ways. First, it could only make good decisions in its deterministic trajectory of expansions for that instance, performing poorly if forced to play from any other state, as has been shown to be possible in the arcade learning environment \cite{revistingarcade}. Second, and maybe more of a concern in our case, an agent could learn a robust strategy that relies on specific characteristics of the training instance and does not generalize well to the larger versions. Step \Step{2} is important to account for the potential diversity of generalization capabilities of the trained agents.

Another idea that might have a positive impact on generalization is stopping training relatively early. This could be useful following the common idea from supervised learning of stopping training when the performance in a testing set starts decreasing. Nevertheless, performance in our case is quite noisy and we have not found strong evidence for that phenomenon being clearly replicated. Another similar but slightly simpler reason to stop training early is that policies might be more diverse during the first stages of training, before convergence is achieved, making the probability of finding a good general strategy there higher.

\subsection{Definition of a Feature Vector}
\label{features}

\newcommand{\specialcell}[2][c]{%
\begin{tabular}[#1]{@{}l@{}}#2\end{tabular}}

\begin{table}[tb]
\centering
\small{
    \begin{tabular}{|c c|}
    \hline
    Feature (size) & Description \\[0.5ex] 
    \hline\hline
    Event label ($|A_{E_p}|$) & Determines the label of $\l$ in $A_{E_p}$. \\
    \hline
    State labels ($|A_{E_p}|$) & \specialcell{Determines the labels of the explored \\ transitions that arrive at $s$.} \\
    \hline
    Controllable (1) & Whether $\l \in A_{E_p}^C$.\\
    \hline
    Marked state (2) & Whether $s$ and $s' \in M_{E_p}$.\\
    \hline
    Phases (3) & \specialcell{Whether (at some point in the episode) \\ a marked state has been found, a \\ winning state has been set, and a cycle \\ containing a marked state was closed. }\\
    \hline
    Child state (3) & \specialcell{Whether $s'$ is winning, losing, none, \\ or not yet explored. }\\
    \hline
    Uncontrollable (4) & \specialcell{Whether $s$ and $s'$ have uncontrollable \\ transitions and they were explored.}\\
    \hline
    Explored (2) & \specialcell{Whether a transition from $s$ or $s'$ has \\ already been explored.}\\
    \hline
    Last expanded (2) & \specialcell{Whether $s$ is the last expanded state \\ in $h$ (outgoing or incoming).}\\
    \hline
    \end{tabular}
\caption{Features that describe a state-action pair $(h, (s, \l))$, where $\D_E(s, \l) = s'$, for a control problem $E_p$ . Size refers to the number of booleans used for each feature.}
\label{tbl:features}
}
\end{table}

The definition of a set of features that compose the state-action signal and describe the transitions in the frontier and the general state of the exploration is a key component of this approach. The feature function $\phi : \S \times \A \into \mathbb{R}^{d_{E_p}}$ should be informative enough to allow good policies. However, it is significantly constrained by the generalization objective. First, the number of features ($d_{E_p}$) should be constant for all instances in a domain, since it is the input dimension of the neural network. Second, the semantics of each element in the feature vector should be maintained and the distribution be shifted as little as possible. A more philosophical constraint is that features should be automatically extracted from $E$, and not defined manually for each domain. A final constraint that is worth mentioning is that they should not be too computationally expensive, since at every step all feature vectors need to be computed and evaluated.

Although real-valued features are possible, they generally rely on the neural network generalizing to unseen values during testing, making generalization harder. The feature vector used in this paper is solely composed of boolean features. The specific features used are shown in Table \ref{tbl:features}. Note that all features are either very inexpensive to compute or are already tracked by \DCS.

A problem with the first two features is that the set $A_{E_p}$ usually depends on the instance parameters $p$. For example, in the Air Traffic problem of the benchmark used \citep{Ciolek:2018:SC-RS-AP}, there is one action label \texttt{land.i} for each plane \texttt{i}, and the number of planes is one of the dimensions $p$. Those indexes need to be removed to address the generalization constraints. In the example, we would only have one label \texttt{land} for feature calculation. This is a significant constraint in our learned exploration policies since they cannot disambiguate different components of the same type.

\section{Experimental Evaluation}
\label{sec:evaluation}

\begin{figure*}[tb]
\centering
\includegraphics[width=\textwidth]{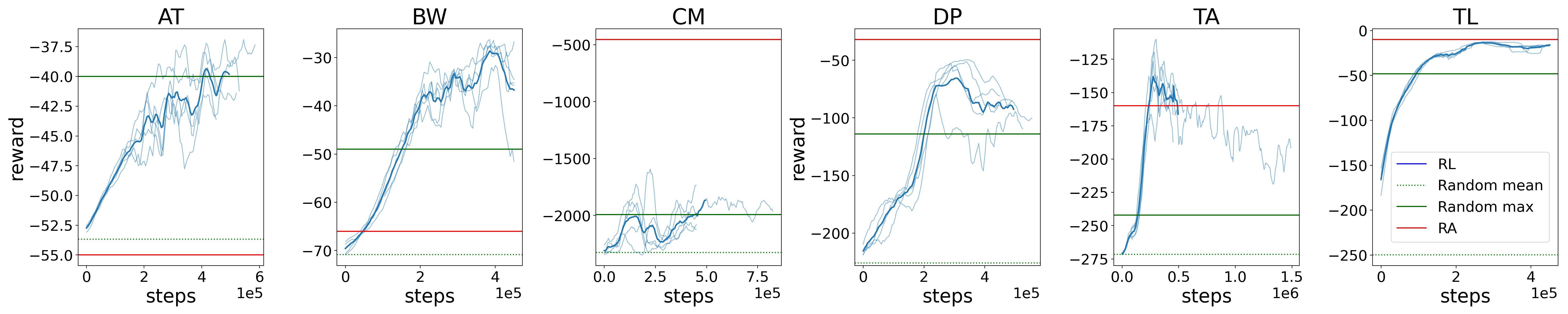}
\caption{Evolution of the number of transitions expanded in the training episodes, with an $\varepsilon$-greedy policy. Results for 5 random seeds are shown, the average highlighted in blue. Curves are smoothed using buckets of $5000$ steps and a moving average of 10 for readability. Performance of \RA and \Random (mean and max over 100 executions) are shown in red and green, respectively.}
\label{fig:training}
\end{figure*}

\begin{figure*}[tb]
    \centering
    \includegraphics[width=\textwidth]{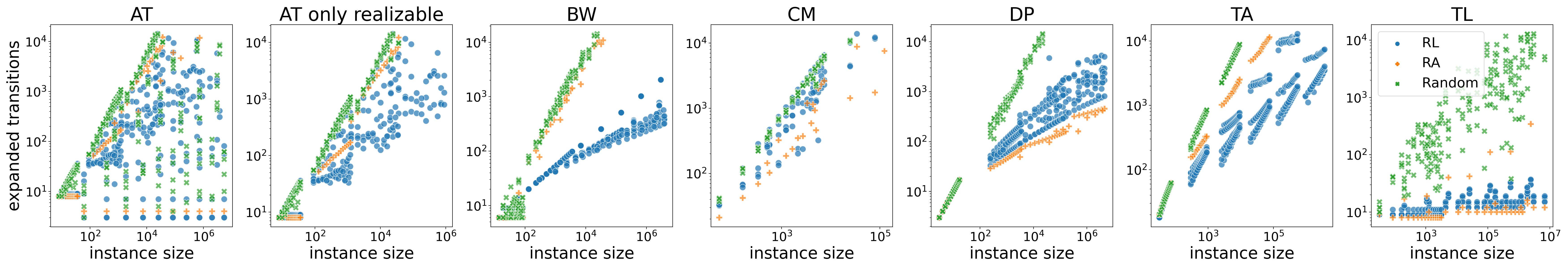}
    \caption{Expanded transitions of the \RA heuristic and each random seed of \RL and \Random, for all instances for which the size of the plant ($x$-axis) could be computed 
    in less that 15000. Second plot is as fist one, but with non-realizable instances removed.}
    \label{fig:scatter generalization}
\end{figure*}

In this section we present empirical results for our approach. We report on an implementation of \DCS extended for feature calculation within the open-source MTSA tool \citep{D'Ippolito:2008:MTSA}. The training procedure, which wraps the synthesis algorithm as an RL environment, is available here\footnote{https://github.com/tdelgado00/Learning-Synthesis}. Experiments were run on an Intel i7-7700 CPU with 16GB of RAM and no GPU. We compare the results with an exploration policy that always chooses a random transition in the frontier (\Random) and with the Ready Abstraction (\RA), the overall best performing heuristic of \citet{CiolekDCS}.

Our approach is evaluated using a benchmark introduced by \citet{Ciolek:2018:SC-RS-AP}. It contains six control problem domains: Air Traffic (AT), Bidding Workflow (BW), Travel Agency (TA), Transfer Line (TL), Dinning Philosophers (DP) and Cat and Mouse (CM). All the problems scale in two dimensions, the number of intervening components grows proportionally to parameter $n$ and the number of states per component grows proportionally to parameter $k$.

For each domain we train in the $(n=2, k=2)$ instance until no better performance is achieved for the last third of the training steps, for a minimum of 500000 steps. The $(2, 2)$ instances range from 91 total transitions in AT to 5044 in CM, so the number of episodes played can vary significantly. While training, we save the weights of the neural network every 5000 steps and a uniform sample of 100 policies is tested with all values of $n$ and $k$ up to 15 with a budget of 5000 transitions \Step{2}, only testing instances ($n$, $k$) for which both $(n-1, k)$ and $(k-1, n)$ have been solved within the budget. After that, we select the neural network that maximizes the number of instances solved, breaking ties with the minimum sum of expanded transitions.

The architecture used is a multilayer perceptron with one hidden layer of 20 neurons and ReLU activation. Informal experimentation showed no improvement using deeper or wider networks, but they might be useful with a larger set of features that allows more complex policies. The optimizer used was stochastic gradient descent with a constant learning rate of $1\text{e-}5$ and weight decay $1\text{e-}4$. The rate of exploration ($\varepsilon$) was decayed linearly from $1.0$ to $0.01$ over the first $250000$ steps of training. The buffer size used for Experience Replay is $10000$, with batch size of $10$, and the target network is reset every $10000$ steps.

Experiments aim to answer the following  questions: 
\begin{itemize}
    \item[\Q{1}] Do agents learn to reduce de exploration of the plant?
    \item[\Q{2}] Are the learned policies competitive in the training instances?
    \item[\Q{3}] Are the policies induced in larger instances competitive?
    \item[\Q{4}] Is the \RL approach competitive with a fixed time budget?
    \item[\Q{5}] Ablation study: What is the impact of the selection step in the overall performance?
\end{itemize}
   
Partial observability, sparse rewards, and the deadly triad are individually sufficient reasons for learning to fail completely, so whether the reward obtained will increase over time is not obvious. Curves in Figure \ref{fig:training} show the evolution of the accumulated reward during training and the performance of \Random for each problem, answering question \Q{1}. First, we highlight that no signs of divergence of the model weights were observed in the results. Furthermore, in all cases non-random average performances were achieved. The learning curves show consistent improvement for AT, BW and TL. In CM we only see a slight improvement with respect to \Random. In DP and TA agents seem to achieve a peek performance that is then lost. Although this loss in performance could be merely explained by the instability of the learning rule, we have observed that removing momentum from the SGD optimizer completely eliminates the decrease. Nevertheless, we chose to keep the momentum since removing it slightly reduces the overall performance of the best agents found.

Even if learning converges, it is initially unclear whether the features chosen are informative enough to encode good policies and whether the agents will be able to find those policies. The red horizontal lines in Figure \ref{fig:training} show the performance of the \RA heuristic, answering question \Q{2}. Our agents rapidly outperform \RA in AT, BW, and TA, and the mean performance in DP and TL approximates that of \RA quite closely. Learning in CM proves to be challenging for our agents, which stay far from the performance of RA.

As discussed in Section \ref{sec:endtoend}, good performances in the training instances do not necessarily translate to larger instances. To evaluate the generalization capabilities of our agents, Figure \ref{fig:scatter generalization} shows for each domain the transitions expanded by the policy selected in Step \Step{2}, \Random and \RA as the total size of the plant grows. Our agents perform significantly better than \Random in all domains, considerably lowering the growth rate of the explored portion of the state space (note that the scale is logarithmic). Furthermore, in AT, BW, and TA, the problems in which training performances were better than \RA, expanded transitions were significantly better in the target instances too. This was not the case in the problems in which our agents did not surpass \RA during training (CM, DP and TL), but in all cases performance was maintained in the larger instances and in DP and TL the agents emulate \RA quite closely. 

Another important remark for these results is the continuity of the expanded transitions across instances for most problems, which is surprising given that different instances are automata with different labels and states, and seems to show that both the set of features and the policies found are being able to capture similarities at least to some degree. In AT all $n > k$ instances are non-realizable (no other benchmark domain has non-realizable instances when $n, k > 1$) making them in most cases easier to solve with few expansions. Note that the behaviour in the realizable instances is more comparable to other problems.

As mentioned in section \ref{sec:algorithm}, our approach entails the overhead of computing features and evaluating the model for every transition in the exploration frontier at each time step, which could be problematic in the end-to-end objective of pushing the frontier of solvable instances within a time budget. 

\begin{table}[tb]
  \centering
  \small{
      \begin{tabular}{lllll}
      \toprule
      & \RL & \RLNS & \Random & \RA \\
      \midrule
      AT & \b{99.2 ± 8.16} & 79.0 ± 3.79 & 82.6 ± 0.8 & 88 \\
      BW & \b{159.4 ± 4.59} & 92.2 ± 25.91 & 47.2 ± 0.4 & 47 \\
      CM & 22.8 ± 0.98 & 23.6 ± 0.8 & 22.0 & 24 \\
      DP & 89.4 ± 11.41 & 62.4 ± 6.62 & 46.6 ± 0.49 & \b{137} \\
      TA & \b{99.2 ± 21.71} & 100.6 ± 22.84 & 51.8 ± 0.4 & 65 \\
      TL & 225.0 & 225.0 & 60.6 ± 4.45 & 225 \\
      All & \b{695.0 ± 23.32} & 582.8 ± 38.01 & 310.8 ± 5.04 & 586 \\
      \bottomrule
      \end{tabular}
  \caption{Number of instances solved by the different approaches.
  Standard deviation is shown when it is not zero.} 
  \label{tbl:solved}
  }
\end{table}

Table \ref{tbl:solved} shows the number of instances solved on average by the selected models and the baselines with a fixed time budget of 10 minutes, answering question \Q{4}. Our approach solves significantly more instances than \RA in three of the six problems (AT, BW, and TA), shows no difference with \RA in two (CM and TL) and solves fewer instances than \RA in DP. Finally, across problems, the total number of instances solved was significantly higher than that of RA. Although in TL both exploration policies solve the same number of instances, if larger instances are evaluated \RA fails first because it performs poorly in the $k=1$ cases (the yellow crosses that grow faster in Figure \ref{fig:scatter generalization}).

Finally, to answer question \Q{5}, the second row (\RLNS) of Table \ref{tbl:solved} shows the number of instances solved when replacing step \Step{2} with a maybe more obvious and cheaper selection method that does not focus on generalization capabilities: selecting the agent with the highest reward in the training instance (breaking ties with the latest agent). Results show that for AT, BW, and DP \Step{2} allowed solving significantly more instances (on average $20.6$, $68.8$, and $23.8$ instances, respectively) while for CM, TA and TL the difference was not significant.
Overall, this step was necessary to solve more instances than \RA.  

\section{Related Work}

Guiding the exploration of the plant in the \DCS algorithm is a Heuristic Search problem in which the objective is to find a subgraph of the plant that contains a winning control strategy (or proves that there is none). A similar approach with RL and generalization has been recently proposed for Classical Planning \citep{planningRL}. A key difference is that they solve a deterministic problem while we study a problem in which the controller does not have full control over the environment. A more comparable setting to DES Control would be Fully Observable Non-Deterministic planning. \citet{planningRL} address reachability properties yielding finite executions, while non-blocking requires infinite executions. In terms of the learning task, their estimated reward is the distance to a goal at a given point in the execution of a plan, while we estimate the number of additional transitions that need to be added to allow \DCS to terminate. 
Additionally, they learn residuals on existing heuristics using reward shaping to accelerate the learning process in what otherwise would be a sparse-reward environment; conversely, we learn from scratch in a sparse-reward environment. Finally, they use Neural Logic Machines to represent the value function, which take logic formulae describing the problem state as input, while we use a traditional multilayer perceptron that takes a general set of features as input.

Our homogeneity hypothesis is similar to the underlying assumption of common patterns in solutions of generalized planning, where different flavors of learning have been applied ~\cite{DBLP:conf/aips/GroshevGTSA18,DBLP:journals/jair/ToyerTTX20,DBLP:conf/kr/StahlbergBG22}.
Plans or policies are represented in such a way they can be applied to solve classical planning problems on any instance of a given domain~\cite{DBLP:journals/apin/MartinG04,DBLP:journals/ai/SrivastavaIZ11}. In our setting, neither exploration policies are algorithmic-like representations \citep[e.g.][]{DBLP:journals/ai/SrivastavaIZ11} nor domain-specific features or lifted domains are defined \citep[e.g.][]{DBLP:conf/kr/StahlbergBG22,DBLP:journals/jair/ToyerTTX20}. In contrast, our approach relies on domain-independent feature representations.

A recent effort in the context of Heuristic Search for classical planning is that of \citet{Sudry_Karpas_2022}.
Despite the differences between DES control and classical planning, they share with our work the key idea of estimating search progress. However, they use supervised learning with LSTMs to estimate relative search progress for a \k{given} heuristic, while our RL approach updates both its estimation of search progress and its search policy simultaneously.

Generalization in RL is an emerging and scarcely studied topic. \citet{kirk2022survey} develop categorizations for tasks and methods for approaching generalization. Following their definitions, we perform out-of-distribution zero-shot policy transfer, training in one context and evaluating in multiple contexts. However, the relation between these MDPs is subtle since they do not share neither states nor actions, and in our approach, this is solved through abstraction. Generalizing to a different set of actions is not considered in their survey and is, to the best of our knowledge, uncommon.

As far as we are aware, DES Control and RL have only been studied jointly by \citet{tatsushi} using tabular RL; but the focus was on the extension to partially observable environments and maximizing permissiveness rather than scaling to larger plants. Their approach requires solving the target control problem repeatedly, whereas ours does not use target instances during training.

\section{Conclusions and Future Work}

In this work we showed a novel way of combining RL and discrete event control, using RL as a heuristic to accelerate a correct and complete control synthesis algorithm.
We proposed a way of framing the guidance of an on-the-fly synthesis algorithm as an RL task and used a modified version of DQN with both states and actions abstracted to make training and generalization feasible. Our results show that learning in small instances is possible; that learned value functions can induce, in larger instances, policies that reduce the explored portion of the plant with respect to human-designed existing heuristics; and that, overall, learned policies can allow solving more control problems within an execution time budget. 
In addition, we highlight a set of components of our approach that can be useful to improve generalization. Namely, selecting exploration policies according to their performances in slightly larger instances, stopping training early to evaluate a diverse set of policies, and defining a set of features that aims at generality rather than completeness.

There is room for improvement in the work described in this paper. We observed by inspecting the specifications that good exploration policies for DP and CM are strongly dependant on the indexes of transition labels (particularly in the latter), which are not informed to our agents due to the generalization restrictions (see Section \ref{features}). We believe that this is the main reason for our agents to underperform in those problems and represents an opportunity for improvement.

More generally, the partial observability of our task could be better addressed. Using a recurrent neural network could allow agents to remember the important aspects of the explored plant and the actions that have been taken. This idea was both proposed for applying DQN on POMDPs \citep{DRQL1} and for estimating search progress for a fixed planning heuristic \citep{Sudry_Karpas_2022}. Nevertheless, it is not immediately adaptable to our setting, where not only states but also actions are partially observable, and all actions are evaluated individually at every step. Alternatively, graph neural networks could be used to process either the explored subgraph or the individual uncomposed automata. However, none of these ideas immediately solve the index problem.

In this paper we train only with one instance of a fixed size for all domains, but a round-robin or incremental training from multiple instances could also be possible. Adding such diversity in the training set has been shown in some cases to reduce overfitting \citep{overfitstudy}, but it is challening in our case since instances in the benchmark used get too large to learn from very rapidly. Additionally, the value function estimates the number of transitions to be expanded and its scale varies with instance size, making it difficult for the learning algorithm to reduce the error towards different instances simultaneously. Similarly, studying generalization across control problems from different domains is of interest. 

Finally, we believe the ideas reported in this paper may be useful for FOND strong cyclic planning and other control settings from the Automated Planning and the Reactive Synthesis communities.

\section*{Acknowledgements}
This work was partially supported by PICT 2018-3835, 2019-1442, 2019-1973; UBACYT 2020-0233BA, 2018-0419BA; and IA-1-2022-1-173516 IDRC-ANII.

\bibliography{root}

\begin{thebibliography}{31}
\providecommand{\natexlab}[1]{#1}

\bibitem[{Camacho, Bienvenu, and
  McIlraith(2021)}]{Camacho_Bienvenu_McIlraith_2021}
Camacho, A.; Bienvenu, M.; and McIlraith, S.~A. 2021.
\newblock Towards a Unified View of AI Planning and Reactive Synthesis.
\newblock \emph{Proceedings of the International Conference on Automated
  Planning and Scheduling}, 29(1): 58--67.

\bibitem[{Ciolek et~al.(2023)Ciolek, Duran, Zanollo, Pazos, Braier, Braberman,
  D’Ippolito, and Uchitel}]{CiolekDCS}
Ciolek, D.; Duran, M.; Zanollo, F.; Pazos, N.; Braier, J.; Braberman, V.;
  D’Ippolito, N.; and Uchitel, S. 2023.
\newblock On-the-fly informed search of non-blocking directed controllers.
\newblock \emph{Automatica}, 147: 110731.

\bibitem[{Ciolek et~al.(2020)Ciolek, Braberman, D’Ippolito, Sardiña, and
  Uchitel}]{Ciolek:2018:SC-RS-AP}
Ciolek, D.~A.; Braberman, V.; D’Ippolito, N.; Sardiña, S.; and Uchitel, S.
  2020.
\newblock Compositional Supervisory Control via Reactive Synthesis and
  Automated Planning.
\newblock \emph{IEEE Transactions on Automatic Control}, 65(8): 3502--3516.

\bibitem[{D'Ippolito et~al.(2008)D'Ippolito, Fischbein, Chechik, and
  Uchitel}]{D'Ippolito:2008:MTSA}
D'Ippolito, N.; Fischbein, D.; Chechik, M.; and Uchitel, S. 2008.
\newblock MTSA: The Modal Transition System Analyser.
\newblock In \emph{Proc. of the Int. Conf. on Automated Software Eng.}, ASE
  '08, 475–476. USA: IEEE Computer Society.

\bibitem[{Ehlers et~al.(2017)Ehlers, Lafortune, Tripakis, and Vardi}]{Ehlers16}
Ehlers, R.; Lafortune, S.; Tripakis, S.; and Vardi, M.~Y. 2017.
\newblock Supervisory control and reactive synthesis: a comparative
  introduction.
\newblock \emph{Discrete Event Dynamic Systems}, 27: 209--260.

\bibitem[{Gehring et~al.(2022)Gehring, Asai, Chitnis, Silver, Kaelbling,
  Sohrabi, and Katz}]{planningRL}
Gehring, C.; Asai, M.; Chitnis, R.; Silver, T.; Kaelbling, L.; Sohrabi, S.; and
  Katz, M. 2022.
\newblock Reinforcement learning for classical planning: Viewing heuristics as
  dense reward generators.
\newblock In \emph{Proc. of the Intl. Conference on Automated Planning and
  Scheduling}, volume~32, 588--596.

\bibitem[{Groshev et~al.(2018)Groshev, Goldstein, Tamar, Srivastava, and
  Abbeel}]{DBLP:conf/aips/GroshevGTSA18}
Groshev, E.; Goldstein, M.; Tamar, A.; Srivastava, S.; and Abbeel, P. 2018.
\newblock Learning Generalized Reactive Policies Using Deep Neural Networks.
\newblock \emph{Proc. of the Intl. Conference on Automated Planning and
  Scheduling}, 28(1): 408--416.

\bibitem[{Hausknecht and Stone(2017)}]{DRQL1}
Hausknecht, M.; and Stone, P. 2017.
\newblock Deep Recurrent Q-Learning for Partially Observable MDPs.
\newblock arXiv:1507.06527.

\bibitem[{Hoffmann et~al.(2020)Hoffmann, Hermanns, Klauck, Steinmetz, Karpas,
  and Magazzeni}]{DBLP:conf/aaai/HoffmannHKSKM20}
Hoffmann, J.; Hermanns, H.; Klauck, M.; Steinmetz, M.; Karpas, E.; and
  Magazzeni, D. 2020.
\newblock Let’s Learn Their Language? A Case for Planning with
  Automata-Network Languages from Model Checking.
\newblock \emph{Proceedings of the AAAI Conference on Artificial Intelligence},
  34(09): 13569--13575.

\bibitem[{Huang and Kumar(2008)}]{Huang:2008:DCD}
Huang, J.; and Kumar, R. 2008.
\newblock Directed control of discrete event systems for safety and
  nonblocking.
\newblock \emph{IEEE Trans. Automation Science \& Engineering}, 5(4): 620--629.

\bibitem[{Kirk et~al.(2022)Kirk, Zhang, Grefenstette, and
  Rocktäschel}]{kirk2022survey}
Kirk, R.; Zhang, A.; Grefenstette, E.; and Rocktäschel, T. 2022.
\newblock A Survey of Generalisation in Deep Reinforcement Learning.
\newblock arXiv:2111.09794.

\bibitem[{Lin(1992)}]{lin1992reinforcement}
Lin, L.-J. 1992.
\newblock \emph{Reinforcement learning for robots using neural networks}.
\newblock Carnegie Mellon University.

\bibitem[{Machado et~al.(2018)Machado, Bellemare, Talvitie, Veness, Hausknecht,
  and Bowling}]{revistingarcade}
Machado, M.~C.; Bellemare, M.~G.; Talvitie, E.; Veness, J.; Hausknecht, M.; and
  Bowling, M. 2018.
\newblock Revisiting the Arcade Learning Environment: Evaluation Protocols and
  Open Problems for General Agents.
\newblock \emph{J. Artif. Int. Res.}, 61(1): 523–562.

\bibitem[{Magee and Kramer(2014)}]{magee2014concurrency}
Magee, J.; and Kramer, J. 2014.
\newblock \emph{Concurrency: State Models and Java Programs}.
\newblock Wiley.

\bibitem[{Mart{\'{\i}}n and Geffner(2004)}]{DBLP:journals/apin/MartinG04}
Mart{\'{\i}}n, M.; and Geffner, H. 2004.
\newblock Learning Generalized Policies from Planning Examples Using Concept
  Languages.
\newblock \emph{Appl. Intell.}, 20(1): 9--19.

\bibitem[{Mnih et~al.(2013)Mnih, Kavukcuoglu, Silver, Graves, Antonoglou,
  Wierstra, and Riedmiller}]{mnih2013playing}
Mnih, V.; Kavukcuoglu, K.; Silver, D.; Graves, A.; Antonoglou, I.; Wierstra,
  D.; and Riedmiller, M. 2013.
\newblock Playing Atari with Deep Reinforcement Learning.
\newblock arXiv:1312.5602.

\bibitem[{Nau, Ghallab, and Traverso(2004)}]{Nau:2004:AP}
Nau, D.; Ghallab, M.; and Traverso, P. 2004.
\newblock \emph{{Automated Planning: Theory \& Practice}}.
\newblock Morgan Kaufmann Publishers.

\bibitem[{Pnueli and Rosner(1989)}]{Pnueli:1989:RS}
Pnueli, A.; and Rosner, R. 1989.
\newblock {On the Synthesis of a Reactive Module}.
\newblock In \emph{Proc. of the Symp. on Principles of Programming Languages},
  POPL, 179--190.

\bibitem[{Ramadge and Wonham(1989)}]{Ramadge:1989:SC}
Ramadge, P.~J.; and Wonham, W.~M. 1989.
\newblock {The control of discrete event systems}.
\newblock \emph{Proc. of the IEEE}, 77.

\bibitem[{Sardi{\~{n}}a and D'Ippolito(2015)}]{DBLP:conf/ijcai/SardinaD15}
Sardi{\~{n}}a, S.; and D'Ippolito, N. 2015.
\newblock Towards Fully Observable Non-Deterministic Planning as
  Assumption-based Automatic Synthesis.
\newblock In \emph{Proc. of the Intl. Joint Conf. on Artificial Intelligence,
  {IJCAI} 2015}, 3200--3206.

\bibitem[{Singh, Jaakkola, and Jordan(1994)}]{QLPOMDP}
Singh, S.~P.; Jaakkola, T.; and Jordan, M.~I. 1994.
\newblock Learning Without State-Estimation in Partially Observable Markovian
  Decision Processes.
\newblock In \emph{Machine Learning Proceedings 1994}, 284--292. Morgan
  Kaufmann.

\bibitem[{Srivastava, Immerman, and
  Zilberstein(2011)}]{DBLP:journals/ai/SrivastavaIZ11}
Srivastava, S.; Immerman, N.; and Zilberstein, S. 2011.
\newblock A new representation and associated algorithms for generalized
  planning.
\newblock \emph{Artif. Intell.}, 175(2): 615--647.

\bibitem[{St{\aa}hlberg, Bonet, and Geffner(2022)}]{DBLP:conf/kr/StahlbergBG22}
St{\aa}hlberg, S.; Bonet, B.; and Geffner, H. 2022.
\newblock Learning Generalized Policies without Supervision Using GNNs.
\newblock In \emph{Procs of the 19th Intl. Conference on Principles of
  Knowledge Representation and Reasoning, {KR}}.

\bibitem[{Sudry and Karpas(2022)}]{Sudry_Karpas_2022}
Sudry, M.; and Karpas, E. 2022.
\newblock Learning to Estimate Search Progress Using Sequence of States.
\newblock \emph{Proceedings of the International Conference on Automated
  Planning and Scheduling}, 32(1): 362--370.

\bibitem[{Sutton and Barto(2018)}]{suttonbarto}
Sutton, R.~S.; and Barto, A.~G. 2018.
\newblock \emph{Reinforcement Learning: An Introduction}.
\newblock Cambridge, MA, USA: A Bradford Book.

\bibitem[{Toyer et~al.(2020)Toyer, Thi{\'{e}}baux, Trevizan, and
  Xie}]{DBLP:journals/jair/ToyerTTX20}
Toyer, S.; Thi{\'{e}}baux, S.; Trevizan, F.~W.; and Xie, L. 2020.
\newblock ASNets: Deep Learning for Generalised Planning.
\newblock \emph{J. Artif. Intell. Res.}, 68: 1--68.

\bibitem[{Ushio and Yamasaki(2003)}]{tatsushi}
Ushio, T.; and Yamasaki, T. 2003.
\newblock Supervisory control of partially observed discrete event systems
  based on a reinforcement learning.
\newblock In \emph{IEEE International Conference on Systems, Man and
  Cybernetics.}, volume~3, 2956--2961. IEEE.

\bibitem[{Watkins and Dayan(1992)}]{Watkins:1992}
Watkins, C. J. C.~H.; and Dayan, P. 1992.
\newblock Technical Note: Q -Learning.
\newblock \emph{Mach. Learn.}, 8(3–4): 279–292.

\bibitem[{Wonham and Ramadge(1987)}]{Wonham:1987:SCS}
Wonham, W.~M.; and Ramadge, P.~J. 1987.
\newblock {On the Supremal Controllable Sublanguage of a Given Language}.
\newblock \emph{SIAM Journal on Control and Optimization}, 25(3).

\bibitem[{Wonham and Ramadge(1988)}]{Wonham:1988:MSC}
Wonham, W.~M.; and Ramadge, P.~J. 1988.
\newblock {Modular Supervisory Control of Discrete-Event Systems}.
\newblock \emph{Mathematics of Control, Signals and Systems}, 1(1): 13--30.

\bibitem[{Zhang et~al.(2018)Zhang, Vinyals, Munos, and Bengio}]{overfitstudy}
Zhang, C.; Vinyals, O.; Munos, R.; and Bengio, S. 2018.
\newblock A Study on Overfitting in Deep Reinforcement Learning.
\newblock arXiv:1804.06893.

\end{thebibliography}

\end{document}